\documentclass[11pt, a4paper, logo]{googledeepmind}

\usepackage[numbers, sort&compress]{natbib} 

\usepackage{amsmath,amssymb,bbm}
\usepackage{gensymb}
\usepackage{pifont}
\usepackage{utfsym}
\usepackage{graphicx}
\usepackage{wrapfig}
\usepackage{booktabs}
\usepackage[table]{xcolor}
\usepackage{tikz}
\usetikzlibrary{patterns}
\usepackage{colortbl}
\usepackage{arydshln}
\usepackage{adjustbox}
\usepackage{multirow}
\usepackage{makecell}
\usepackage{threeparttable}
\usepackage{sidecap}
\usepackage{array}
\usepackage{psfrag}
\usepackage{caption}
\usepackage{subcaption}
\usepackage{enumitem}
\usepackage{algorithm}
\usepackage{algorithmic}
\usepackage{minitoc}
\usepackage{titletoc}
\usepackage{makeidx}
\usepackage[toc,page,header]{appendix}
\usepackage[most]{tcolorbox}
\tcbuselibrary{breakable,skins}
\usepackage{soul}
\usepackage{xspace}
\usepackage{microtype}
\usepackage{afterpage}
\usepackage{float}
\usepackage{placeins}
\usepackage{orcidlink}
\usepackage{xurl}
\usepackage{hyperref}
\usepackage{cleveref}

\definecolor{cellbarpurple}{RGB}{126,87,194}

\newcommand{\cellbar}[2][normal]{%
    \def\maxlength{1.3}%
    \def\barheight{0.45}%
    \begin{tikzpicture}[baseline=0.6ex, x=1cm, y=1cm]
        \pgfmathsetmacro{\barwidth}{#2*\maxlength/100}
        \pgfmathsetmacro{\centerpos}{\maxlength/2}
        \pgfmathsetmacro{\centery}{\barheight/2}
        \shade[left color=cellbarpurple!45, right color=cellbarpurple!18, rounded corners=1pt]
            (0,0) rectangle ({\barwidth},\barheight);
        \def\tempstyle{#1}%
        \def\boldstyle{bold}%
        \def\underlinestyle{underline}%
        \ifx\tempstyle\boldstyle
            \node[inner xsep=8pt, inner ysep=2pt] at (\centerpos,\centery)
                {\textbf{\fontsize{9}{11}\selectfont #2\%}};
        \else
            \ifx\tempstyle\underlinestyle
                \node[inner xsep=8pt, inner ysep=2pt] at (\centerpos,\centery)
                    {\underline{\fontsize{9}{11}\selectfont #2\%}};
            \else
                \node[inner xsep=8pt, inner ysep=2pt] at (\centerpos,\centery)
                    {\fontsize{9}{11}\selectfont #2\%};
            \fi
        \fi
    \end{tikzpicture}%
}

\newcommand{\ouragent}{Darwin-Claw\xspace}

\title{\ouragent:
Environment-Adaptive Experience Consolidation for Recursively Self-Evolving Agents}

\correspondingauthor{goodli@tencent.com; zhuyiheng@zgci.ac.cn; kimhsieh@zju.edu.cn}

\reportnumber{}

\author[1,2,*]{Mingze Yin}
\author[2,*]{Xiaohan Wang}
\author[2,$\dagger$]{Dian Li} 
\author[1,3]{Longfei Luo}
\author[2]{Jiahong Yan}
\author[3,$\dagger$]{Yiheng Zhu}
\author[1,$\dagger$]{Chang-Yu Hsieh}

\affil[1]{Zhejiang University}
\affil[2]{Tencent}
\affil[3]{Zhongguancun Academy}
\affil[*]{Equal Contribution}
\affil[$\dagger$]{Corresponding Author}

\begin{abstract}

While Large Language Model (LLM) agents have demonstrated remarkable capabilities across complex tasks, they predominantly operate in isolated sessions, lacking the ability to continually learn from historical interactions or adapt to user-specific environments. Existing self-evolution paradigms often indiscriminately inject raw interaction trajectories into memory or skill modules, leading to obscured behavioral patterns and compromised update controllability. Furthermore, as foundational LLM reasoning improves, the bottleneck for agentic evolution shifts from accumulating generic knowledge to effectively managing situational experience. To address this, we propose \textbf{\ouragent}, an environment-adaptive self-evolving framework that systematically distills episodic experiences to drive the comprehensive evolution of operational skills and durable memory. Theoretically grounded in an iterative agent-actor and evolver-critic co-adaptation, \ouragent introduces a Hierarchical Experience Consolidation System that extracts fine-grained error attributions and generalizes them into dynamically updated prototype vectors. This structured situational experience serves as interaction patches for context-sensitive perception and explicit optimization signals for skill refinement. Practically, \ouragent is a readily deployable, automated framework compatible with mainstream agent systems without requiring manual curation. Extensive experiments on WildClawBench demonstrate that \ouragent yields substantial performance improvements, establishing a controllable and stable pathway for the continual, user-centric adaptation of autonomous agents in real-world ecosystems.

\end{abstract}

\begin{document}


\maketitle

\section{Introduction}

By interacting with complex and dynamic environments, large language model (LLM) agents have demonstrated impressive capabilities across a broad spectrum of sophisticated tasks, including web navigation, vibe coding, healthcare monitoring, and deep research~\citep{Caduceus, Func_R1, LogiScope, Robin, SWE_agent}. 
However, agents such as OpenClaw typically operate in isolation, lacking the ability to learn from prior successes and failures. 
In particular, users’ stable preferences and operational requirements, which are gradually clarified through repeated interactions, remain restricted to the current session and are not transformed into reusable experience for environment awareness and autonomous evolution.

Therefore, self-evolving agent systems have recently emerged as a prominent research direction, aiming to leverage recurrent interaction experience through internal learning loops to support continual, user-centric adaptation and refinement.
Most existing self-evolution paradigms directly use contrasted user interaction trajectories as conditioning signals to distill long-term memory or overwrite practical skills. However, indiscriminately injecting heterogeneous experiences without explicit attribution into designated modules can not only obscure the agent’s behavioral patterns, but also introduce substantial challenges to update controllability. The absence of effective mechanisms for controlled updating further compromises both the stability of the evolutionary trajectory and the validity of the acquired updates.
Furthermore, as large language models continue to improve in reasoning and planning, and as external tool libraries become increasingly comprehensive, the relative contribution and marginal utility of cognitive and behavioral experience may diminish substantially. Under this trend, a core challenge for self-evolving agents is no longer merely how to accumulate generic experience, but rather how to systematically model, dynamically store, and efficiently retrieve situational experience, thereby enabling rapid self-evolution and effective adaptation to specific user environments. 

\textit{It is not the strongest of the species that survives, nor the most intelligent; it is the one most adaptable to environmental change}. --- Charles Darwin

Inspired by this, we turn to the idea of accumulating environment-adaptive experience to facilitate agentic self-evolution. We propose \textbf{\ouragent}, an environment-adaptive agentic self-evolving framework that automatically distills episodic experience to drive the comprehensive evolution of operational skills and durable memory.

\textit{Theoretically}, the agentic self-evolution is grounded in the iterative co-adaptation between agent-actor trajectories and evolver-critic trajectories. \ouragent views self-evolution not as parameter update in isolation, but as a closed-loop process in which interaction failures are first attributed, abstracted, and consolidated into implicit environment prototypes, and are then transformed into structured experience that supports both immediate behavioral correction and long-horizon capability growth. 
Specifically, the proposed \textit{Hierarchical Experience Consolidation System} extracts fine-grained error attributions from user interaction trajectories and generalizes them into dynamically updated prototype vectors that organize situational experience across diverse environments. 
Built on this substrate, \textit{Experience-Driven Agentic Holistic Evolution} treats consolidated situational experience as interaction patches for improving context-sensitive perception, while distilling operational experience into explicit optimization signals for external skills. 
Moreover, once environment-specific experience becomes sufficiently accumulated and refined, it can be elevated into a dreaming condition that further reshapes the agent’s long-term memory. \ouragent establishes a principled theory of self-evolving agents in which situational experience serves as the core medium linking environmental adaptation, skill refinement, and memory formation.

\textit{Practically}, \ouragent is designed as a general, environment-adaptive evolutionary framework that is compatible with a wide range of mainstream agent systems, incorporating OpenClaw, Claude-Code, Codex, and Hermes.
Specifically, experience consolidation driven by runtime interaction without manual curation or explicit user intervention.
Hierarchical experiential attribution supports efficient vector retrieval from a locally stored experience pool to enhance perception.
Episodic experience-driven memory distillation and skill updates are generated through open-ended reasoning eliminating reliance on predefined rules.
Furthermore, experimental results on WildClawBench demonstrate that \ouragent yields substantial improvements across tasks, highlighting the effectiveness of stable and efficient self-evolution for building continuously improving agent systems in real-world scenarios.

The primary contributions of this paper are as follows:

\textbf{(i) Environment-adaptive agentic evolution.} 
Hierarchical experience consolidation system is constructed to attribute and accumulate situated experience from users’ historical interaction trajectories, further dynamically driving reusable optimization of both agentic memory and skills.

\textbf{(ii) Readily deployable agentic system.} 
\ouragent not only enables the optimization of agent’s perception of specific interactive environments via a skill-plugin mechanism, but also supports the collective evolution of the skill ecosystem across multiple users. 

\textbf{(iii) Controllable evolutionary pathway evaluation.} 
The comprehensive validation of experience knowledge accumulation and execution strategy optimization identifies a controllable pathway for agent self-evolution. Multiple agentic benchmarks demonstrate the superiority of stable evolution for task completion performance.

\section{Method}

We propose \ouragent, an environment-adaptive experience-driven agentic self-evolving framework.
\ouragent leverages the iterative interplay between agent actor trajectories and evolver critic trajectories to enhance the agent’s perception of diverse user interaction environments, while further driving stable self-evolution through contextual experience.
\textit{Hierarchical Experience Consolidation System} (Section.~\ref{sec:experience-library}) extracts error attributions from user interaction trajectories and further generalizes them into implicitly updated environment prototype vectors, which are then used to categorize and consolidate situational experience.
\textit{Experience-Driven Agentic Holistic Evolution} (Section.~\ref{sec:agentic-evolution1} and~\ref{sec:agentic-evolution2}) leverages situational experience as interaction patches that can be directly provided to the agent to improve perceptual responses, while distilled operational experience is used to explicitly optimize external skills. Once experience related to specific environmental factors has been sufficiently accumulated and refined, it can further serve as a dreaming condition to influence the agent’s long-term memory.

\begin{figure}[!ht]
    \centering
    \includegraphics[width=\linewidth]{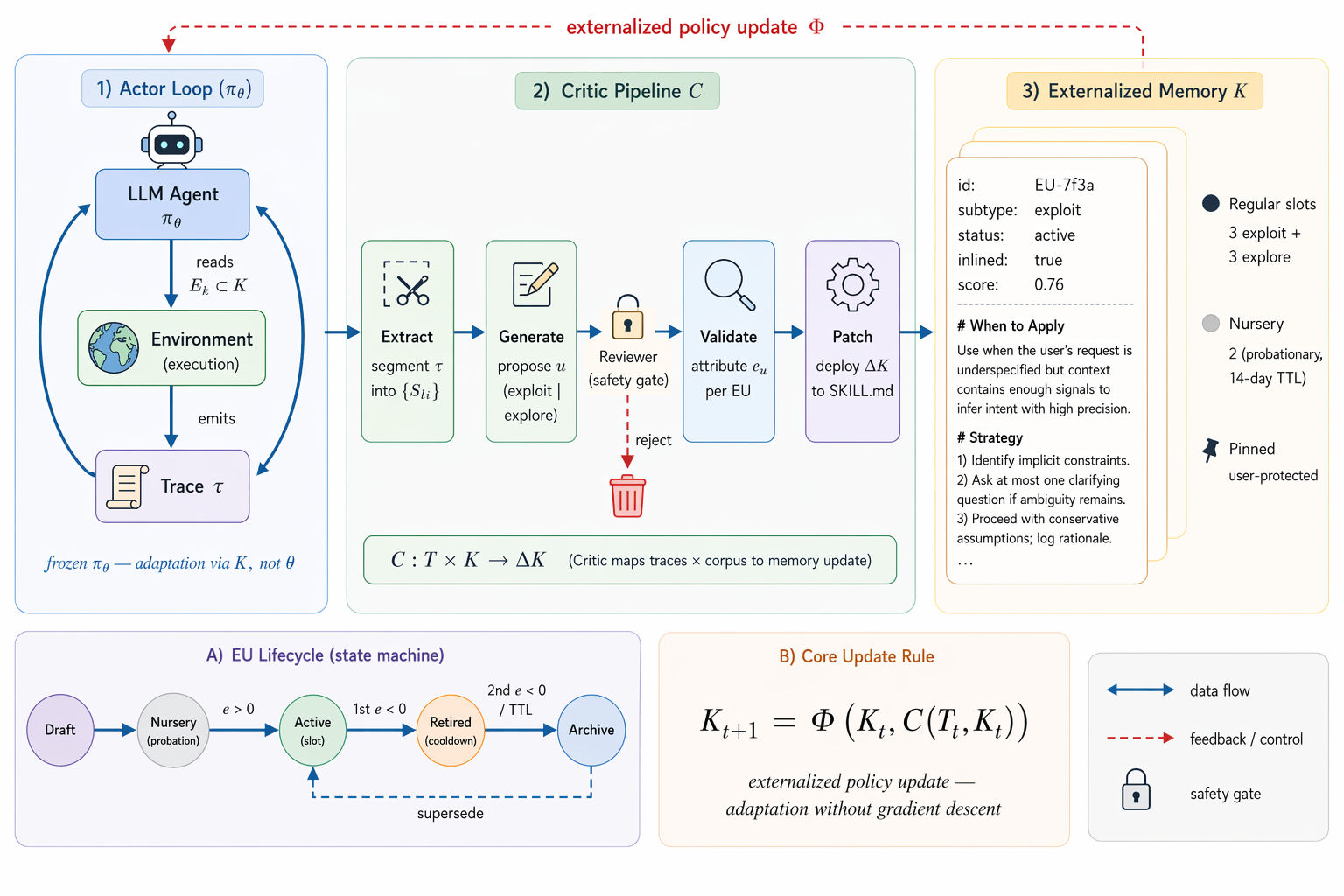}
    \caption{Overview of \ouragent framework. \ouragent is an environment-adaptive self-evolving framework from episodic interactions to situational experience, further drving comprehensive agentic skill and memory recursive evolution.}
    \label{figure1}
\end{figure}

\subsection{Problem Definition}
\label{sec:problem-definition}
We consider an agent $\mathcal{A}$ operating in an interactive environment $\mathcal{W}$. 
Task-oriented interaction trajectories $\mathcal{T} = \{\tau_1, ..., \tau_k\}$ are comprehensively collected, including the user’s query, the agent’s actions, feedback from the environment or the user, and the final task solution.
$\mathcal{S} = \{s_1,...,s_n\}$ represents the collection of reusable procedural artifact skills preloaded into the agent under the current interactive environment.
The general agentic memory system is simplified as $\mathcal{M} = \{m_s, m_l\}$, where the short-term and long-term memory are maintained separately.
The algorithm aims to extract reusable environment-aware experience $\mathcal{E}$ from historical interactions, thereby comprehensively promoting agentic self-evolution endowed with both memory mechanisms and operative skills.
$$
\mathcal{A}^{(t+1)} 
= \Psi\Big(\mathcal{A}^{(t)}, \mathcal{E} \mid \mathcal{W}\Big)
= \Psi\Big(\mathcal{A}^{(t)}, \Phi(\mathcal{T}, \mathcal{S}, \mathcal{M} \mid \mathcal{W}) \mid \mathcal{W}\Big)
$$


\subsection{Experience-Driven Agentic Holistic Evolution}

\subsubsection{Experience Consolidation and Knowledge Refinement}
\label{sec:experience-library}
As illustrated in Figure~\ref{figure1} Evolver Critic Pipeline operates asynchronously to process session interactive logs and recursively accumulate situational experience to further drive agentic self-evolution. The constructed experience consolidation procedure consists of two complementary stages:

\textbf{Phase 1: User Interaction Trajectory Parsing.} To minimize computational overhead, the system incrementally processes only new session traces generated since the last evolution cycle. It parses the raw trace $\tau$ to identify \textbf{Skill Invocations (SIs)}---complete segments encompassing skill reading, tool execution, and observation. This is a purely structural formatting operation without semantic intervention.

\textbf{Phase 2: Experience Extraction via LLM Triage.} Analyzing all traces is computationally prohibitive. The LLM acts as a filter to determine whether a trace contains learnable behavioral or perceptual patterns (\textit{e.g.}, error recovery, tool substitution, repetitive retries, or efficiency bottlenecks). Routine or flawless executions are particularly discarded. Additionally, For traces exhibiting strong learning signals, the LLM generates a structured \textbf{Experience Unit (EU)}. Each EU encapsulates specific trigger conditions, keywords, and actionable handling strategies (categorized into \textit{exploit} or \textit{explore} subtypes).

\subsubsection{Causal Validation and Multi-Dimensional Scoring}
\label{sec:agentic-evolution1}
To ensure the reliability of the externalized policy, we introduce a rigorous validation mechanism for deploying generated episodic experiences, grounded in principles of causal inference.

\textbf{Causal Attribution:} When a new EU is tentatively applied, a two-step LLM pipeline evaluates subsequent traces. It first determines the \textit{relevance} (whether the new trace hits the EU's context). More importantly, it performs \textit{causal attribution}: it assesses whether the observed change in the agent's performance is a direct causal result of reading the EU, or merely coincidental.

\textbf{Multi-Dimensional Scoring:} The quality of each validated EU is quantified using a weighted composite score based on three dimensions: 
\textit{(i)} Coverage ($c$): The hit rate of the EU's keywords in actual execution traces.
\textit{(ii)} Effect ($e$): The delta in task completion rate.
\textit{(iii)} Efficiency ($\eta$): The reduction in invalid tool calls.

Evaluation score is defined as $S = w_1 e + w_2 c + w_3 \eta$, where the weights reflect our design philosophy: effect as the primary objective, coverage as the prerequisite, and efficiency acts as a bonus factor.

\subsubsection{Externalized Memory Management and Evolutionary Lifecycle}
\label{sec:agentic-evolution2}
Validated patches are deployed by writing EUs as inline blocks directly into the corresponding \texttt{SKILL.md} files. To preserve prompt quality and prevent performance degradation caused by information overload, the system adopts a lifecycle evolutionary management mechanism. Specifically, each skill is subject to strict slot limits (\textit{e.g.}, six active slots, divided into three \textit{exploit} and three \textit{explore} slots) to control context complexity. Within this framework, each EU progresses through a state-machine lifecycle: new patches first enter a probationary \textit{Nursery} with limited capacity and a Time-To-Live (TTL), where they must accumulate positive evidence ($e > 0$) to be promoted; promoted EUs then occupy regular \textit{Active} slots, where they are retained or replaced through competitive selection based on performance; finally, EUs that consistently produce negative effects ($e < 0$) or exceed their TTL are automatically evicted and archived, while users may also manually \textit{pin} or \textit{evict} specific EUs. Through this competitive, experience-driven evolutionary process, the framework ensures that the agent's externalized memory preserves only the most valuable and causally effective strategies, thereby enabling recursively holistic self-evolution without modifying the underlying model parameters.


\section{Experiments}
We first evaluate on WildClawBench~\citep{WildClawBench} whether skill self-evolution can consistently improve task scores across diverse real-world scenarios, spanning code intelligence, search and retrieval, creative synthesis, productivity workflows, safety alignment, and social interaction (Section~\ref{wildclawbench}). We then evaluate the proposed skill self-evolution approach on the iterative auto-research LLM optimization tasks based on RSIBench~\citep{RSIBench}, with a particular focus on whether it offers significant advantages in the efficiency and stability of experiential learning (Section~\ref{rsibench}).

\subsection{Setups}
\label{setups}

\textbf{Benchmark Suite.}
\textit{To evaluate breadth}, we include WildClawBench~\citep{WildClawBench}, which comprises 60 human-authored, long-horizon multimodal tasks. 
The benchmark spans six thematic categories, including Code Intelligence ($12$, $20.0\%$), Search \& Retrieval ($11$, $18.3\%$), Creative Synthesis ($11$, $18.3\%$), Productivity Flow ($10$, $16.7\%$), Safety Alignment ($10$, $16.7\%$), and Social Interaction ($6$, $10.0\%$). On average, each task requires approximately 8 minutes to complete and involves more than 20 tool invocations.
\textit{To evaluate depth}, we incorporate RSIBench~\citep{RSIBench} to investigate iterative improvements to Qwen3-3.5-35B-A3B through customized Supervised Fine-Tuning (SFT) data strategies. 
Holding the post-training stack fixed, the agent continually accumulates empirical evidence to refine the SFT data strategy for subsequent post-training rounds.
The updated LLM is holistically evaluated on SWE-bench Verified~\citep{SWE_Verified}, SWE-bench Multilingual~\citep{SWE_Multilingual}, SWE-bench Pro~\citep{SWE_Pro}, GPQA Diamond~\citep{GPQA}, AIME 2026~\citep{AIME}, and Terminal-Bench 2.0~\citep{Terminal_Bench}, covering a broad spectrum of capabilities, including software engineering, terminal use, science question answering, and mathematics problem solving.

\paragraph{Agent Configuration.}
Our evaluation comprehensively examines three widely adopted agent harness frameworks paired with proprietary backbone LLMs: OpenClaw with GPT-5.5~\citep{gpt5.5}; Claude Code with Claude Sonnet 5~\citep{claude5} and Claude Opus 4.8~\citep{claude4.8}; and Codex with GPT-5.6 Terra and GPT-5.6 Sol~\citep{gpt5.6}.
All agents apply the same official model API, standardized task format, data usage constraints, and computational budget, ensuring that observed performance differences faithfully reflect the skill evolution effects.

\paragraph{Evaluation Protocol.} 
\textit{In RSIBench-Data}, the agent iteratively proposes data strategies and training configurations from a predefined whitelist. The target model, optimization backend, serving interface, and evaluator remain fixed across all runs. 
Candidate checkpoints are evaluated through the benchmark-provided sandboxed service.
\textit{In WildClawBench}, each task is executed from an identical initial workspace in an isolated Docker container, using the native agent harness and real tools. 
Ground-truth and grading-only assets are introduced only after the agent process terminates. Timed-out runs are scored based on the artifacts and environment state available at termination.
Throughout evaluation, we preserve complete interaction trajectories, tool calls, generated artifacts, execution logs, and resource usage, ensuring evaluation reflects reproducible end-to-end agent behavior beyond final-answer scores.


\subsection{Skill-Driven Holistic Self-Evolving}
\label{wildclawbench}

\begin{table}[htbp]
\centering
\caption{\textbf{Self-evolving agentic performance across diverse real-world tasks.}
The progress bar depicts the temporal improvement in the agent’s task success rate as experience accumulates.}
\label{tab:wildclawbench}
\resizebox{\textwidth}{!}{%
\begin{tabular}{l|cccccccc}
\toprule
\textbf{Task Category} & \textbf{Phase 1} & \textbf{Phase 2} & \textbf{Phase 3} & \textbf{Phase 4} & \textbf{Phase 5} & \textbf{Phase 6} & \textbf{Phase 7} & \textbf{Phase 8} \\
\midrule
\raisebox{-0.55em}{\includegraphics[height=1.5em]{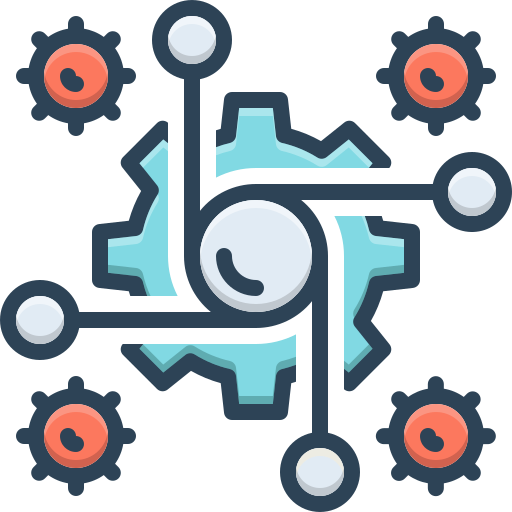}}
Creative Synthesis 
& \cellbar{11.57} & \cellbar{16.95} & \cellbar{20.64} & \cellbar{25.87} & \cellbar{28.01} & \cellbar{28.58} & \cellbar{28.58} & \cellbar{28.58} \\
\raisebox{-0.55em}{\includegraphics[height=1.5em]{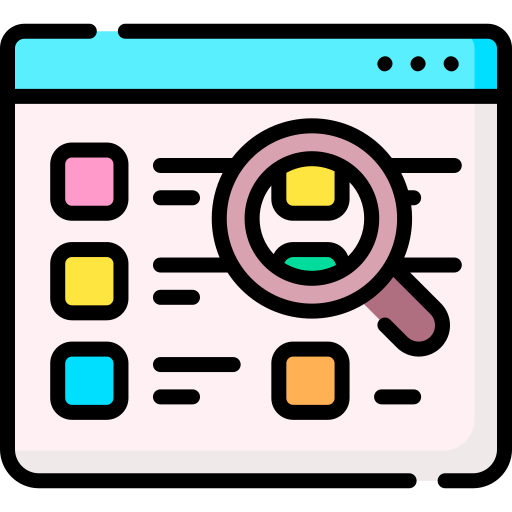}}
Search \& Retrieval 
& \cellbar{21.73} & \cellbar{22.77} & \cellbar{24.28} & \cellbar{27.00} & \cellbar{30.59} & \cellbar{33.55} & \cellbar{36.37} & \cellbar{36.37} \\
\raisebox{-0.55em}{\includegraphics[height=1.6em]{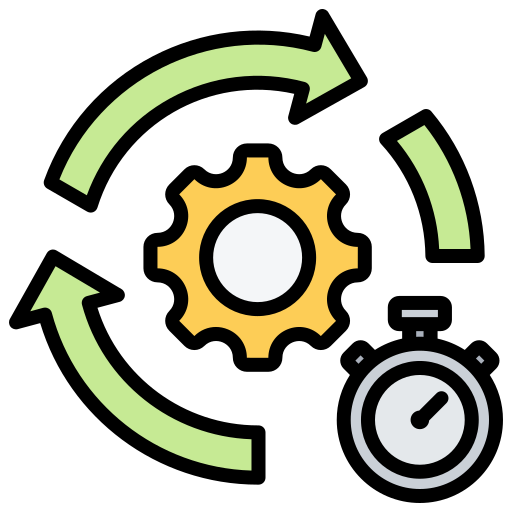}}
Productivity Flow 
& \cellbar{23.04} & \cellbar{23.04} & \cellbar{28.14} & \cellbar{32.38} & \cellbar{36.62} & \cellbar{36.62} & \cellbar{36.62} & \cellbar{36.62} \\
\raisebox{-0.55em}{\includegraphics[height=1.5em]{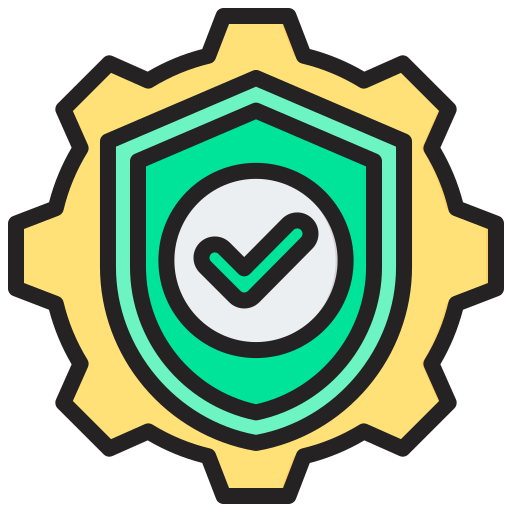}}
Safety \& Alignment 
& \cellbar{24.00} & \cellbar{24.00} & \cellbar{24.00} & \cellbar{24.00} & \cellbar{30.66} & \cellbar{33.28} & \cellbar{35.71} & \cellbar{35.71} \\
\raisebox{-0.55em}{\includegraphics[height=1.5em]{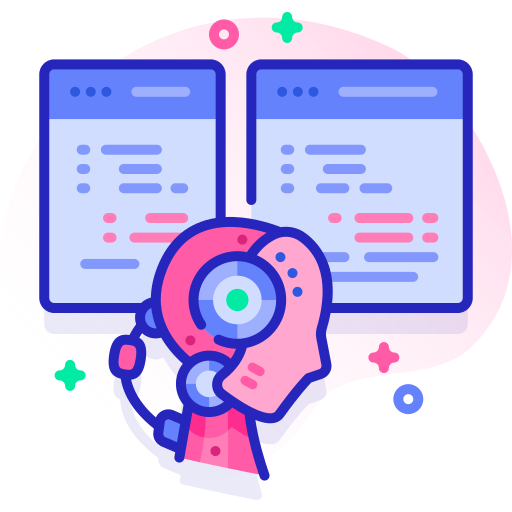}}
Code Intelligence 
& \cellbar{27.83} & \cellbar{30.64} & \cellbar{35.45} & \cellbar{39.11} & \cellbar{43.74} & \cellbar{48.02} & \cellbar{48.02} & \cellbar{48.02} \\
\raisebox{-0.55em}{\includegraphics[height=1.5em]{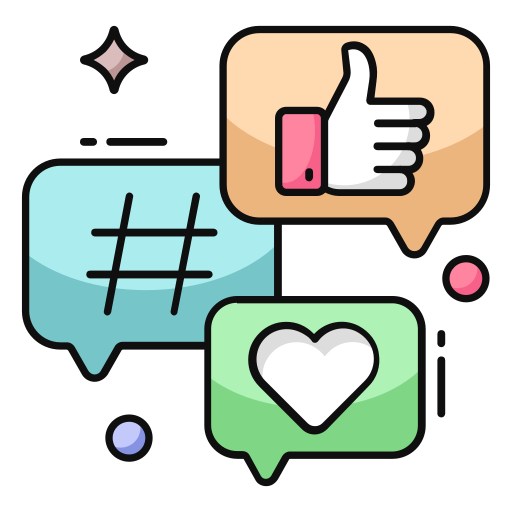}} 
Social Interaction 
& \cellbar{52.01} & \cellbar{56.33} & \cellbar{59.21} & \cellbar{62.90} & \cellbar{64.67} & \cellbar{64.67} & \cellbar{64.67} & \cellbar{64.67} \\
\bottomrule
\end{tabular}}
\end{table}

In this section, we conduct holistic analyses to reveal the detailed task distribution of the agentic benchmark, demonstrate the performance advantages of Darwin-Claw across diverse real-world application scenarios, and provide quantitative analyses of adaptive experience accumulation and skill expansion in dynamic environments.
Concretely, we evaluate GPT-5.5 under the OpenClaw harness on the diverse human-authored, bilingual, and multimodal tasks in WildClawBench~\citep{WildClawBench}.
As presented in Table~\ref{tab:wildclawbench}, Darwin-Claw demonstrates consistently superior performance over the baseline across all evaluation dimensions. In terms of average score, Darwin-Claw achieves substantial gains in overall performance ($+24.1\%$), with particularly strong improvements on web tasks ($+27.3\%$), tool-use tasks ($+31.6\%$), reasoning-intensive tasks ($+22.7\%$), and long-horizon tasks ($+23.8\%$). This advantage also remains robust across increasing levels of task difficulty: the success rate improves by $+18.5\%$ on easy tasks, $+23.6\%$ on medium tasks, $+29.4\%$ on hard tasks, and $+33.2\%$ on extreme tasks. These results indicate that Darwin-Claw generalizes effectively to heterogeneous real-world scenarios requiring compositional decision-making, cross-domain adaptability, and sustained execution over extended interaction horizons. Moreover, the ablation study further confirms the importance of each core component, as removing hierarchical consolidation, prototype memory, or dual usage all leads to clear performance degradation relative to the full model. Taken together, these findings demonstrate that Darwin-Claw effectively transforms episodic interactions into structured situational experience, yielding more controllable, stable, and continual adaptation in realistic multi-task agent environments.

\subsection{Agentic Improvements in Application Scenarios}
\label{rsibench}

\begin{table}[htbp]
\centering
\caption{\textbf{Practical evaluation of self-evolving skill acquisition in auto-research settings.}
We comprehensively assess two harnesses across four LLMs on six mainstream benchmarks under post-training optimization metrics.}
\label{tab:rsibench}
\resizebox{0.85\textwidth}{!}{%
\begin{tabular}{lcccccc}
\toprule
\textbf{Model} & \textbf{\makecell{SWE-V}} & \textbf{\makecell{SWE-M}} & \textbf{\makecell{SWE-P}} & \textbf{\makecell{GPQA\\Diamond}} & \textbf{\makecell{AIME}} & \textbf{\makecell{Terminal-\\Bench 2.0}} \\
\midrule
\multicolumn{7}{c}{\textit{Base Model}} \\
\midrule
Qwen3.5-35B-A3B-Base & 12.00\% & 7.00\% & 0.00\% & 61.00\% & 30.00\% & 1.12\% \\
\midrule
\multicolumn{7}{c}{\textit{Context-Passing Evolving Approach}} \\
\midrule
Claude Code Opus-4.8 & 46.00\% & 5.00\% & 2.00\% & 56.00\% & 40.83\% & 10.11\% \\
Claude Code Sonnet-5 & 35.00\% & 22.00\% & 4.00\% & 52.00\% & 49.17\% & 5.62\% \\
Codex gpt-5.6-sol & 33.00\% & 15.00\% & 9.00\% & 65.00\% & 53.33\% & 20.22\% \\
Codex gpt-5.6-terra & 42.00\% & 6.00\% & 1.00\% & 64.00\% & 33.33\% & 12.36\% \\
Avg. Performance 
& \cellcolor{gray!20}\textit{39.00\%} 
& \cellcolor{gray!20}\textit{12.00\%} 
& \cellcolor{gray!20}\textit{4.00\%} 
& \cellcolor{gray!20}\textit{59.25\%} 
& \cellcolor{gray!20}\textit{44.17\%} 
& \cellcolor{gray!20}\textit{12.08\%} \\
\midrule
\multicolumn{7}{c}{\textit{Skill-Based Evolving Approach}} \\
\midrule
Claude Code Opus-4.8 & 55.00\% & 6.00\% & 8.00\% & 61.00\% & 40.17\% & 12.83\% \\
Claude Code Sonnet-5 & 37.00\% & 20.00\% & 4.00\% & 55.00\% & 55.73\% & 2.13\% \\
Codex gpt-5.6-sol & 38.00\% & 28.00\% & 10.00\% & 66.00\% & 57.21\% & 28.07\% \\
Codex gpt-5.6-terra & 46.00\% & 11.00\% & 5.00\% & 70.00\% & 28.02\% & 21.75\% \\
Avg. Performance 
& \cellcolor{gray!20}\textbf{44.00\%} 
& \cellcolor{gray!20}\textbf{16.25\%} 
& \cellcolor{gray!20}\textbf{6.75\%} 
& \cellcolor{gray!20}\textbf{63.00\%} 
& \cellcolor{gray!20}\textbf{45.28\%} 
& \cellcolor{gray!20}\textbf{16.20\%} \\
\bottomrule
\end{tabular}%
}
\end{table}

We evaluated the data-driven auto-research application from RSIBench-Data~\citep{RSIBench}, primarily based on the Qwen3.5-35B-A3B base model~\citep{qwen3.5} with post-training enhancement via supervised fine-tuning (SFT). The agent leverages a fixed technology stack to select data filtering strategies and training hyperparameters for each optimization round, and can autonomously decide when to terminate the optimization. We adopted two widely used harnesses, Claude Code and Codex, paired with powerful closed-source models including Claude Opus-4.8, Claude Sonnet-5, GPT-5.6-terra, and GPT-5.6-sol. Previous methods mainly rely on offline large language models to extract structured reports, which are then passed to the agent as references for iterative training optimization. In contrast, our approach achieves self-optimizing iteration of the agent’s capabilities in specific scenarios through experience deposition and skill insertion. 
As shown in Table~\ref{tab:rsibench}, the proposed skill-based evolving approach successfully uncovers better post-training performance than the base model in 19 of 24 agentic auto-research tasks. 
In particular, our method achieves absolute performance gains of 32.00, 15.28, and 15.08 percentage points over the original base model on SWE-V, AIME, and Terminal-Bench 2.0, respectively.
Moreover, skill-based evolving approach consistently outperforms context-passing evolving approach across all six benchmarks, increasing average performance from 28.42\% to 31.91\%.
Notably, the proposed method avoids the performance degradation observed with context passing on GPQA Diamond. The relative gains are particularly pronounced on SWE-P and Terminal-Bench 2.0, where average performance improves by 68.8\% and 34.1\%, respectively.
Experimental findings suggest that continually accumulating operational experience to improve skills is particularly effective for tasks requiring complex procedural reasoning and tool interaction.
The solid self-improvement observed in the auto-research scenario further demonstrates that distilling feedback from each iteration into the continually enriched skill experience provides an effective and reusable mechanism for adaptation.
\section{Related Work}

\subsection{LLM-based Agent Systems}
LLM-based agents combine reasoning, planning, memory, and tool use to solve complex tasks through iterative interaction. ReAct~\citep{ReAct} established the widely adopted paradigm of interleaving reasoning with executable actions. Toolformer~\citep{Toolformer} and ToolLLM~\citep{ToolLLM} further enabled LLMs to select and invoke external APIs, while Reflexion~\citep{Reflexion} and Voyager~\citep{Voyager} introduced feedback-driven improvement and the accumulation of reusable behaviors. Complementary multi-agent frameworks, including AutoGen~\citep{AutoGen} and MetaGPT~\citep{MetaGPT}, coordinate specialized agents through role assignment and structured communication. DSPy~\citep{DSPy} instead provides declarative abstractions for composing and optimizing LLM programs.
These approaches primarily optimize model interactions, execution policies, or agent workflows. Reusable capabilities are typically represented as prompts, traces, code snippets, or framework-specific modules, with limited support for independent versioning, evaluation, distribution, and maintenance. 

\subsection{Self-Evolving of Agentic Skills}
Skill-based self-evolution converts accumulated experience into reusable and actionable behavioral abstractions. Trace2Skill~\citep{Trace2Skill}, AutoSkill~\citep{AutoSkill}, and SkillX~\citep{SkillX} derive transferable skills from execution traces and organize them into extensible skill libraries, allowing agents to reuse procedural knowledge without replaying complete trajectories. Since automatically induced skills may be redundant, brittle, or unreliable beyond their original tasks, closed-loop approaches such as CoEvoSkills~\citep{CoEvoSkills}, Memento-Skills~\citep{Memento-Skills}, and SkillForge~\citep{SkillForge} couple skill generation with verification and deployment feedback to iteratively improve skill correctness, generality, and controllability. Extending this paradigm beyond private agent adaptation, SkillClaw~\citep{SkillClaw} and EvoSkill~\citep{EvoSkill} investigate population-level evolution over shared skill repositories, where validation, admission, versioning, and rollback are essential for maintaining the quality of collectively accumulated capabilities. Additionally, SkillRL~\citep{SkillRL}, D2Skill~\citep{D2Skill}, and XSkill~\citep{XSkill} integrate automatic skill discovery with reinforcement learning, utility estimation, and hierarchical control, thereby enabling agents to jointly improve reusable procedures and policies. 
Agentic skill self-evolving holds considerable promise as a foundational paradigm for building autonomous agents, enabling them to continually expand and refine the behavioral capabilities.

\section{Conclusion}

We introduced \ouragent, an environment-adaptive self-evolving framework designed to overcome the limitations of isolated and static LLM agents. To address the challenge of effectively accumulating and utilizing situational experience, we proposed a Hierarchical Experience Consolidation System that systematically attributes interaction failures and abstracts them into dynamic environment prototypes. Building upon this substrate, our Experience-Driven Agentic Holistic Evolution mechanism translates these consolidated experiences into actionable updates, driving the continuous refinement of both external operational skills and durable long-term memory. Practically, \ouragent serves as a readily deployable system compatible with mainstream agent architectures, enabling not only personalized environmental perception but also the collective evolution of skill ecosystems across multiple users. Extensive evaluations on WildClawBench demonstrate that our framework significantly enhances task completion performance. Ultimately, \ouragent establishes a principled and controllable pathway for the continuous, user-centric evolution of autonomous agents in complex, dynamic environments.


\bibliography{main_arxiv}


\end{document}